\documentclass[runningheads]{llncs}

\usepackage{eccv}

\usepackage{eccvabbrv}

\usepackage[table]{xcolor}
\usepackage{graphicx}
\usepackage{booktabs}
\usepackage{amsmath,amssymb}
\usepackage{color}
\usepackage{kotex}
\usepackage{xcolor}
\usepackage{caption}
\newcommand{\km}[1]{\textcolor{black}{#1}}

\usepackage{multirow}
\usepackage{wrapfig}
\usepackage{array}  
\usepackage{algorithm}
\usepackage{graphicx}
\usepackage{xcolor}
\usepackage{amsmath}
\usepackage{algpseudocode}
\usepackage{wrapfig}
\newcommand{\hf}{\raisebox{-0.15em}{\includegraphics[height=1em]{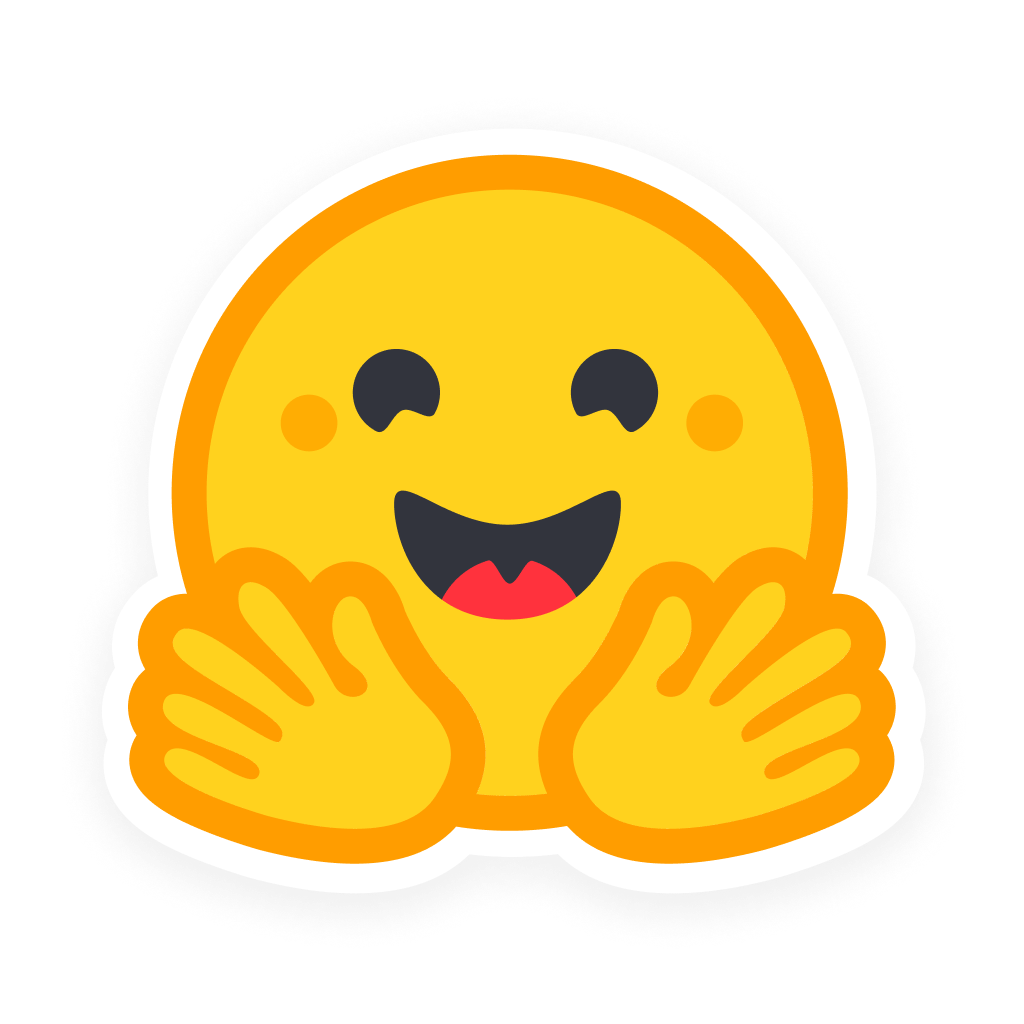}}}
\usepackage[accsupp]{axessibility}  

\usepackage[hyperfootnotes=false]{hyperref}

\usepackage{orcidlink}

\begin{document}

\title{Zero-Shot Quantization for Object Detectors using Off-the-Shelf Generative Models}

\titlerunning{\textbf{G}enerative \textbf{o}ff-the-shelf models for \textbf{o}bject \textbf{d}etector \textbf{Q}uantization}




\author{Hyunho Lee$^{1,\ast}$\orcidlink{0009-0003-2544-4994} \and
Kyomin Hwang$^{1,\ast}$\orcidlink{0009-0004-9364-7950} \and
Hyeonjin Kim$^{1,\ast}$\orcidlink{0009-0005-8001-1387} \and
Suyoung Kim$^{1}$\orcidlink{0009-0002-1925-6201} \and
Sunghyun Wee$^{1,2}$\orcidlink{0009-0008-9898-2570} \and
Nojun Kwak$^{1,\dagger}$\orcidlink{v}}
\institute{$^{1}$Seoul National University, Seoul, South Korea \\ $^{2}$LG Electronics, Seoul, South Korea\\[3pt]
\email{\{hhlee822, kyomin98, peaceful1, ksyo96, wsh05, nojunk\}@snu.ac.kr}}

\authorrunning{H.~Lee et al.}


\maketitle

\renewcommand\thefootnote{}
\footnotetext{$^{\dagger}$\,Corresponding author.}
\footnotetext{$^{\ast}$\,Equal contribution.}
\renewcommand\thefootnote{\arabic{footnote}}

\begin{abstract}
With an increasing number of Object Detection (OD) models being deployed on edge devices, Zero-Shot Quantization for OD (ZSQ-OD) aims to quantize these models when access to the original training data is prohibited. Existing research on Zero-Shot Quantization-Aware Training (QAT) for OD synthesizes training sets through noise optimization. However, this approach struggles to maintain performance in low-bit regions. In this paper, we introduce \textbf{GoodQ} (\textbf{G}enerative \textbf{o}ff-the-shelf models for \textbf{o}bject \textbf{d}etector \textbf{Q}uantization), a QAT pipeline that utilizes off-the-shelf generative models to construct a training set. We first identify three challenges that arise when introducing a generative model to the ZSQ-OD task: 1) each image contains dense information with multiple instances, 2) the class-wise distribution in the original dataset is imbalanced, and 3) the pseudo-labels assigned to the generated images can potentially act as noisy signals during QAT. GoodQ addresses these challenges by 1) introducing an Information-Dense Prompting strategy to generate multi-instance images, 2) applying Intrinsic Distribution-Aware Selection to match the pretrained class distribution, and 3) employing Teacher-guided Adaptive Noise Reduction to mitigate noise arising from the QAT process. Our framework achieves state-of-the-art performance in low-bit ZSQ (W4A4) and extends quantization to extreme bit-widths (W3A3). Furthermore, we conduct an extensive analysis to uncover the underlying factors contributing to the efficacy of GoodQ. The constructed dataset is available at \href{https://huggingface.co/datasets/Kyomin/GoodQ}{\hf~\texttt{GoodQ}}.

\end{abstract}

\section{Introduction}

\begin{figure*}[t]
    \centering
    \includegraphics[width=1.0\linewidth]{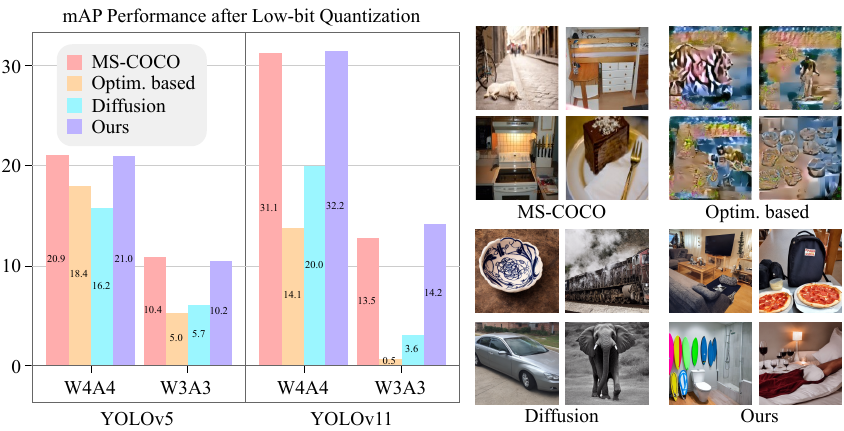} 
    \caption{\textbf{\textsc{Left}} Performance degradation across different quantization bit-widths for the optimization-based method (Optim. based) and our proposed approach, relative to the real dataset baseline. \textbf{\textsc{Right}} A visual comparison of images from the real dataset alongside those generated with optimization, diffusion, and our method.}
\label{fig:teaser}
\end{figure*}

Quantization has become a standard practice for deploying deep learning models in resource-constrained environments. Quantization-Aware Training (QAT) effectively compresses full-precision models to lower bit-widths by fine-tuning them on a small training set selected from the original training dataset. However, access to the original data is often infeasible in real-world applications due to privacy regulations. To address this, Zero-Shot Quantization (ZSQ) has emerged as a promising paradigm~\cite{zeroq,intraq,hast,qimera,synq,mimiq}. ZSQ synthesizes a set of proxy samples to substitute for the original dataset during training. While early ZSQ methods were based on noise optimization relying on batch normalization (BN) statistic matching~\cite{zeroq}, subsequent research has shifted toward enhancing the diversity of these synthetic samples to better approximate the distribution of the original dataset~\cite{hast,qimera,enhancingDFQ}. Notably, recent works such as GenQ~\cite{genq} and MixQ~\cite{mixup_genq} have demonstrated that integrating off-the-shelf generative models is an effective approach for diversifying synthetic training sets in classification tasks.

As in image classification, we hypothesize that data diversity is crucial for maintaining quantization quality in Object Detection (OD) tasks. To verify this hypothesis, we conducted a preliminary experiment. Using the OD-specific ZSQ algorithm introduced in TSOD~\cite{zsqod}, we compared three dataset configurations: a subset of the real MS-COCO dataset~\cite{coco_data}, the noise-optimized synthetic dataset proposed in TSOD, and a diffusion-generated dataset\footnote{Following GenQ, we used the prompt, ``A photo of a $\{\text{D}\}$ $\{\text{C}\}$,'' where $\{\text{D}\}$ is a CLIP ImageNet template and $\{\text{C}\}$ is a randomly sampled class label.}. Figure~\ref{fig:teaser} illustrates these results by showing the mAP scores of the three configurations. As shown in the figure, the noise-optimized synthetic dataset exhibits a limitation: failure in low-bit regions. While the noise-optimized configuration suffers an excessive performance drop, the diverse dataset generated by an off-the-shelf diffusion model achieves relatively better quantization performance at low-bit region. This observation supports our hypothesis that promoting dataset diversity is important in OD, motivating us to leverage diffusion models to enhance quantization quality.

Based on the aforementioned observation, this paper aims to achieve further improvements by addressing three challenges specific to OD: 1) Unlike image classification, OD requires each image to contain dense semantic information, often encompassing multiple object categories. 2) The class-wise distribution in OD datasets is inherently imbalanced, making it difficult to accurately reproduce these distributional characteristics. 3) When conducting QAT with training data synthesized by generative models, the pseudo-labels assigned to these images may introduce noisy supervision signals, potentially misleading the QAT process.

To address these challenges, we propose \textbf{GoodQ} (\textbf{G}enerative \textbf{o}ff-the-shelf models for \textbf{o}bject \textbf{d}etector \textbf{Q}uantization), a pipeline that leverages off-the-shelf generative models tailored specifically for the ZSQ-OD task. GoodQ tackles each challenge through 1) Information-Dense Prompting, 2) Intrinsic Distribution-Aware Selection, and 3) Teacher-guided Adaptive Noise Reduction. These components enable GoodQ to effectively utilize off-the-shelf generative models within the context of ZSQ-OD. To evaluate the effectiveness of the proposed method, we conduct extensive experiments on YOLO-based OD models~\cite{yolov11, yolov5}, demonstrating that GoodQ excels in low-bit regions while maintaining competitive performance at higher bit-widths. Beyond quantitative results, we also provide comprehensive analyses to explain the effectiveness of GoodQ.

In summary, our main contributions are as follows:
\begin{itemize}
    \item We investigate the utilization of off-the-shelf generative models for the ZSQ-OD task and identify the key challenges associated with this approach.
    \item We propose GoodQ, a novel framework designed to tailor generative models specifically for OD tasks, effectively mitigating the identified challenges.
    \item Through extensive experiments and analysis, we demonstrate the underlying factors contributing to the effectiveness of GoodQ.
\end{itemize}
\section{Related Works}

\subsection{Quantization}

Quantization reduces the memory footprint and enables efficient inference by mapping high-precision tensors to low-bit integers. Quantization-Aware Training (QAT)~\cite{lsq_plus,qkd,lsq,fcn} utilizes a small training set to determine quantization parameters and update model weights. Specifically, the weights are first mapped to integers using initial quantization parameters and then fine-tuned on the curated training set. The simulated quantization of a high-precision tensor $x$ to a $b$-bit representation is formulated as follows:

\begin{equation}
\label{eq:quantization}
    \hat{x} = s \times \left\lfloor\mathrm{clip}\!\left(\frac{x}{s}, \, n, \, p \right)\right\rceil,
\end{equation}

\noindent where $n=-2^{b-1}$ and $p=2^{b-1}-1$ for symmetric weight quantization, while $n=0$ and $p=2^b-1$ for asymmetric activation quantization. Here, $s$ is a learnable scaling parameter, $\lfloor\cdot\rceil$ denotes the rounding operation, and $\mathrm{clip}(\cdot)$ is a function that clamps values to the range $[n, p]$.

\subsection{Zero-Shot Quantization} 

While the majority of QAT methods curate their training sets from the original dataset, there are cases where access to these data is restricted due to privacy regulations and security concerns. To address this limitation, researchers have investigated quantization methods that employ synthetic datasets. This line of research is referred to as Zero-Shot Quantization (ZSQ)~\cite{genq,synq,zeroq,mixup_genq,clampvit}.

Synthetic training dataset for ZSQ is primarily generated in three ways: 1) through noise optimization, 2) by training a GAN-based generator, and 3) by leveraging an off-the-shelf (mostly diffusion-based) generative model. Noise optimization methods utilize the pretrained model to directly optimize learnable inputs initialized with Gaussian noise~\cite{synq,zeroq}. GAN-based methods, on the other hand, train an adversarial network so that the generator can accurately approximate the internal knowledge of the model~\cite{gdfq,ris,enhancingDFQ}. However, both noise optimization and GAN-based methods struggle to maintain performance when quantized to lower bit-widths due to a lack of data diversity. To tackle this issue, methods that leverage off-the-shelf generative models have emerged~\cite{genq,mixup_genq,genvit}, achieving state-of-the-art performance.

\subsection{Zero-shot Quantization for Object Detection} 

However, most previous research on ZSQ has primarily focused on image classification, largely overlooking other tasks such as Object Detection (OD)---a task in high demand for deployment on edge devices in real-world scenarios. TSOD~\cite{zsqod} is a pioneering work that addresses ZSQ specifically for OD (ZSQ-OD) by incorporating a task-specific QAT objective, $\mathcal{L}_\text{detect}$, alongside the task-agnostic term, $\mathcal{L}_\text{distill}$. It proposes a noise-optimization-based method to synthesize a training set by leveraging task-specific information, achieving superior performance. Despite this, TSOD exhibits a critical limitation: it struggles to maintain robustness in low-bit quantization regions (e.g., W4A4). This performance degradation at low bit-widths hinders its applicability when a quantized model needs to be deployed in resource-constrained environments.

\subsection{Position of Our Work}

As empirically demonstrated in Figure~\ref{fig:teaser}, the performance degradation of TSOD in low-bit regions can potentially be mitigated by utilizing synthetic training sets generated by off-the-shelf models. Building on this observation, we propose a ZSQ-OD pipeline that leverages such generative models in a manner tailored specifically for OD tasks. To achieve this, we first identify three key challenges that must be addressed. By systematically resolving these challenges, we formulate our comprehensive ZSQ-OD pipeline: GoodQ.
\section{Three Challenges of ZSQ-OD} \label{sec:observation}

In this section, we introduce three challenges that must be addressed when adapting off-the-shelf generative models specifically for the OD task: two arising during the training set generation stage, and one during the Quantization-Aware Training (QAT) stage. These challenges can be summarized as follows:

\begin{enumerate}
\item \textbf{Information Density Challenge}: A single image may contain multiple instances, and thus multiple labels.
\item \textbf{Class-wise Imbalance Challenge}: Each class may contain a different number of instances due to the multi-label nature of OD datasets.
\item \textbf{Pseudo-label Challenge}: The pseudo-labels assigned to each image can introduce noisy supervision signals during the QAT process.
\end{enumerate}

\subsection{Information Density Challenge}

A primary characteristic of OD datasets is the high density of information present within each image. For instance, the MS-COCO 2017 dataset~\cite{coco_data}---one of the most widely used OD datasets---contains an average of 7.7 bounding boxes (BBoxes) per image. While object sparsity has not been a critical issue for single-label tasks such as image classification, OD is a multi-label task that inherently depends on both categorical information and spatial localization (e.g., BBoxes). Consequently, there is a compelling need to generate images with information-rich content to adequately satisfy the requirements of OD tasks.

\subsection{Class-wise Imbalance Challenge}

Another challenge arises from the extreme class-wise distribution skew inherent in OD datasets. For instance, in the full MS-COCO dataset, the proportions of bounding boxes for the most frequent (head) and least frequent (tail) classes are roughly 30.52\% and 0.02\%, respectively. This represents a far more severe long-tail imbalance than is typically observed in image classification tasks. Consequently, a naive application of generative models for image generation is likely to overlook or misrepresent this heavily skewed distribution.

\subsection{Pseudo-Label Challenge}

The final challenge arises from the pseudo-labels assigned to the generated images. The objective of QAT is to guide the initially quantized model to match the behavior of the full-precision model using a small training set. The detection-specific term of the ZSQ-OD objective ($\mathcal{L}_\text{detect}$) takes two inputs: $\mathbf{y}$, which is the target serving as guidance, and $\hat{\mathbf{y}}$, which is the output of the quantized model being optimized. The objective is formulated as follows:

{\small
\begin{equation}
\label{eq:l_detect}
    \mathcal{L}_\text{detect}(\hat{\mathbf{y}}, \mathbf{y}) = \lambda_\text{b} \underbrace{\left(1 - \text{CIoU}(\hat{\mathbf{y}}_\text{b}, \mathbf{y}_\text{b})\right)}_{\mathcal{L}_\text{box}} + \lambda_\text{o} \underbrace{\text{BCE}\bigl(\hat{y}_\text{o},\, y_\text{o}\bigr)}_{\mathcal{L}_\text{obj}} + \lambda_\text{c} \underbrace{\text{BCE}\bigl(\hat{\mathbf{y}}_\text{c},\,\mathbf{y}_\text{c}\bigr)}_{\mathcal{L}_\text{cls}}.
\end{equation}
}

TSOD employs $\mathbf{y}^\text{GT} \triangleq(\mathbf{y}^\text{GT}_{b},y^\text{GT}_{o}, \mathbf{y}^\text{GT}_{c})$ as $\mathbf{y}$, where $\mathbf{y}^\text{GT}_{b} \in \mathbb{R}^{4}, \ y^\text{GT}_{o} \in \mathbb{R}$, and $\mathbf{y}^\text{GT}_{c} \in \mathbb{R}^{|\mathcal{C}|}$ represent the ground truths for bounding box coordinates, objectness score, and class probabilities, respectively. However, when utilizing an off-the-shelf generative model, the generated images are unlabeled, requiring a pseudo-labeling process to substitute for $\mathbf{y}^\text{GT}$. Because the pseudo-labeling process generally enforces one-hot style hard labels, it introduces potential noise that inherently depends on the performance of the model utilized for labeling.
\section{Method}

\begin{figure*}[t] 
    \centering
    \includegraphics[width=1.0\linewidth]{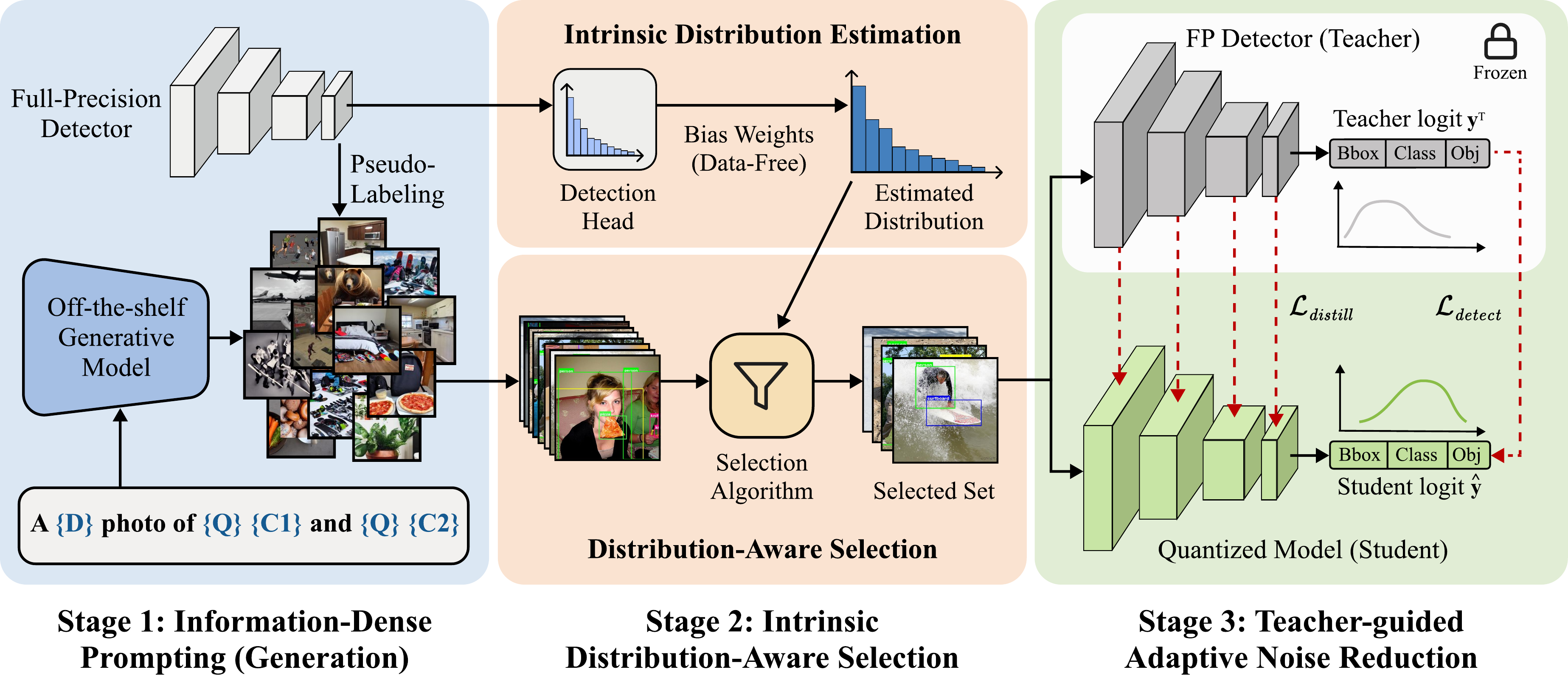}     
\caption{Overview of the GoodQ pipeline. \textbf{\textsc{Stage 1}} Information-Dense Prompting constructs an image pool tailored for Object Detection using an off-the-shelf generative model. \textbf{\textsc{Stage 2}} Intrinsic Distribution-Aware Selection estimates the inherent class distribution and selects a representative training dataset from the image pool. \textbf{\textsc{Stage 3}} Teacher-guided Adaptive Noise Reduction utilizes teacher-generated soft predictions as targets to mitigate label noise during the QAT process.}
\label{fig:overview} 
\end{figure*}

To address the aforementioned challenges, we propose GoodQ, a ZSQ-OD pipeline that leverages off-the-shelf generative models. Our method consists of two primary steps: 1) constructing a training dataset tailored specifically for QAT in OD, and 2) performing the QAT process using the constructed dataset. Figure~\ref{fig:overview} illustrates the overview of our proposed method. In the following sections, we provide a detailed explanation of our approach across three stages.

\subsection{Stage 1. Information-Dense Prompting}
\label{sec:IDP}

Unlike image classification, ZSQ-OD inherently necessitates information-dense images where multiple object categories and BBoxes coexist within a single scene. To address the Information Density Challenge, we propose Information-Dense Prompting (IDP) to better reflect the characteristics of real-world OD datasets. IDP employs the following template:

\begin{align}
\text{Prompt} = \text{``A \{D\} photo of \{Q\} \{C1\} and \{Q\} \{C2\}''},
\end{align}

\noindent where $\{C_1\}$ and $\{C_2\}$ denote class labels, $\{D\}$ represents a descriptive adjective (e.g., clean, nice), and $\{Q\}$ is a quantitative adjective (e.g., multiple, many). A detailed list of the prompts used in our experiments is provided in the Appendix. By leveraging IDP, we generate a pool of 160k candidate images with a high density of BBoxes tailored for ZSQ-OD. Subsequently, we employed a pretrained detection model to generate bounding box annotations, thereby assigning pseudo-labels to the synthesized images.
\begin{algorithm}[t]
\caption{Distribution-Aware Selection}
\label{alg:greedy_selection__}
\begin{algorithmic}[1]
\Require Image pool $\mathcal{I}$, target size $K$, target ratios $\{\hat{r}_{c}\}$, BBox counts $B_{i, c}$, ${c \in \mathcal{C}}$
\Ensure Selected image set $\mathcal{S}$
\State $\mathcal{S} \gets \emptyset$
\While{$|\mathcal{S}| < K$ and $\mathcal{I} \neq \emptyset$}
    \Statex \textbf{\#Step 1: Target Distribution Estimation \& Candidate Update}
    \State Compute current ratios $\{p_c\}$ from current $\mathcal{S}$
    \For {$c \in \mathcal{C}$}
        \State Compute gaps $g_c \gets \hat{r}_c - p_c$ \Comment{$g_c > 0$ for under-selected class $c$}
        \State $\mathcal{A}_c \gets \{ i \in \mathcal{I} \mid B_{i, c} > 0 \}$ \Comment{Set of every image with class $c$}
    \EndFor
    \Statex \textbf{\#Step 2: Priority-Based Target Class Selection}
    \State $\mathcal{C}_{\mathrm{select}} \gets \{ c \in \mathcal{C} \mid g_c > 0 \text{ and } \mathcal{A}_c \neq \emptyset \}$ \Comment{Get under-selected class}
    \If{$\mathcal{C}_{\mathrm{select}} \neq \emptyset$}
        \State $c^* \gets \arg\min_{c \in \mathcal{C}_{\mathrm{select}}} |\mathcal{A}_c|$ \Comment{least remaining class}
    \Else
        \State $c^* \gets \arg\max_{c:\, \mathcal{A}_c \neq \emptyset} g_c$ \Comment{least over-selected class}
    \EndIf
    \Statex \textbf{\#Step 3: BBox-Driven Image Selection \& Update}
    \State $i^* \gets \arg\max_{i \in \mathcal{A}_{c^*}} \sum_{c \in \mathcal{C}} B_{i,c}$ \Comment{Choose $i^*$ with the most BBox}
    \State $\mathcal{S} \gets \mathcal{S} \cup \{i^*\}$, \quad $\mathcal{I} \gets \mathcal{I} \setminus \{i^*\}$ \Comment{Move $i^*$ to selected image set}
\EndWhile
\State \textbf{return} $\mathcal{S}$ (pad randomly from $\mathcal{I}$ if $|\mathcal{S}| < K$)
\end{algorithmic}
\end{algorithm}

\subsection{Stage 2. Intrinsic Distribution-Aware Selection}

OD datasets exhibit a more inherently skewed class distribution compared to image classification. To address the Class-wise Imbalance Challenge, we propose Intrinsic Distribution-Aware Selection (IDAS). IDAS aligns the distribution of the generated images with that of the original dataset through two steps: 1) estimating the distribution of the real dataset in a data-free setting, where the original training images and labels are entirely inaccessible, and 2) selecting images from the generated pool to match this estimated class-wise distribution.

\subsubsection{Intrinsic Distribution Estimation}

In data-free scenarios, access to the class distribution of the original dataset is strictly restricted. Consequently, the distribution must be estimated from the only available source: the pre-trained model weights. The bias terms of the detector may implicitly capture information about the class prior. Therefore, to reliably estimate the target distribution from this limited information source, we leverage the bias parameters of the detection head. Specifically, the target class-wise ratio is computed as follows:

\begin{equation}
\label{eq:r_c}
\hat{w}_c = \bar{b_c} - \min_{i \in \mathcal{C}} \bar{b_i}, \quad
\hat{r}_c = \frac{\max\!\left(\hat{w}_c / \sum_{k \in \mathcal{C}}\hat{w}_k,\ \epsilon\right)}
{\sum_{j \in \mathcal{C}}\max\!\left(\hat{w}_{j} / \sum_{k \in \mathcal{C}}\hat{w}_k,\ \epsilon\right)}, 
\end{equation}

\noindent where $\bar{b}_{c}$ denotes the average bias value for class $c$ extracted from the pre-trained detection head, $\mathcal{C}$ is the set of all object classes, and $\epsilon$ is a small constant serving as a correction factor to prevent zero target ratios during normalization. By applying Equation~(\ref{eq:r_c}), we derive the set of target ratios $\mathcal{R} \triangleq \{ \hat{r}_{c} \mid c \in \mathcal{C} \}$. Notably, the distribution estimated via our approach exhibits a strong Pearson correlation ($r = 0.87$) with the ground-truth class distribution of the MS-COCO dataset. This high correlation provides empirical validation for our proposed estimation strategy and demonstrates its effectiveness in accurately capturing the target distribution even in a data-free setting.

\subsubsection{Distribution-Aware Selection}

Leveraging the pre-computed target ratios $\mathcal{R}$ from the previous section, we introduce Distribution-Aware Selection (DAS) to curate the generated image pool to better reflect the estimated target distribution. Given the pool of images generated in Section~\ref{sec:IDP}, we perform a greedy selection process until the set of selected images $\mathcal{S}$ reaches the target size $K$.The selection process follows these steps in each iteration: we calculate the class distribution $p_c$ of the images in $\mathcal{S}$ and compute the discrepancy between $p_c$ and $\hat{r}_c$ (\textsc{Step 1}). We then rank the classes by prioritizing those with the largest discrepancy, while incorporating an additional preference for those with fewer remaining candidate images (\textsc{Step 2}). Based on this ranking, a target class is chosen. Among the images that contain the chosen class, we select the one with the largest number of detected BBoxes and update $\mathcal{S}$ (\textsc{Step 3}). The complete procedure is detailed in Algorithm~\ref{alg:greedy_selection__}.

\subsection{Stage 3. Teacher-guided Adaptive Noise Reduction}

Assigning pseudo-labels to diffusion-generated images using a pre-trained detection model inevitably introduces label noise. Therefore, it is inappropriate to directly apply the conventional Object Detection QAT loss $\mathcal{L}_\text{detect}$ in Equation (\ref{eq:l_detect}) to our setting. To mitigate the adverse effects of such noisy supervision, Teacher-guided Adaptive Noise Reduction (TANR) bypasses pseudo-labels entirely by constructing $\mathcal{L}_\text{detect}$ directly from the teacher's soft predictions as targets. Specifically, we replace the pseudo-label with the soft label $\mathbf{y}^{\mathcal{T}}\triangleq(\mathbf{y}^\mathcal{T}_{b},y^\mathcal{T}_{o}, \mathbf{y}^\mathcal{T}_{c})$ produced by the full-precision teacher model. Furthermore, we adopt QFocal-based Adaptive Weighting (AW)~\cite{qfocal}, which yields the revised loss:

\begin{equation}   
\label{eq:aw}
\mathcal{L}_{\text{detect}}(\hat{\mathbf{y}}, \mathbf{y}^{\mathcal{T}})=\lambda_\text{b}\mathcal{L}_{\text{box}}+|y^\mathcal{T}_{\text{o}}-\hat{y}_{\text{o}}|^\gamma \lambda_\text{o}\mathcal{L}_{\text{obj}}+\sum_{\text{c}_i\in \mathcal{C}} |y^\mathcal{T}_{\text{c}_i}-\hat{y}_{\text{c}_i}|^\gamma \lambda_\text{c}\mathcal{L}_{\text{cls}},
\end{equation}
\noindent where $\gamma$ denotes a scaling factor. $\mathbf{y}_c^\mathcal{T}$ and $\hat{\mathbf{y}}_c$ denote the teacher's and student's class-probability vectors over the class set $\mathcal{C}$, respectively. For each class $c_i \in \mathcal{C}$, $y_{c_i}^\mathcal{T}$ and $\hat{y}_{c_i}$ denote the $i$-th elements of $\mathbf{y}_c^\mathcal{T}$ and $\hat{\mathbf{y}}_c$, corresponding to the predicted probability of class $c_i$. AW guides the QAT process by enabling the student model to focus on instances that are more susceptible to quantization-induced noise. Finally, we combine our $\mathcal{L}_\text{detect}$ with the conventional $\mathcal{L}_\text{distill}$ to formulate the overall QAT objective.
\section{Experiments}

\begin{table*}[t]
\centering
\caption{Quantization Aware Training (QAT) results on YOLOv5/YOLOv11 using MS-COCO validation set. We fix the size of the calibration set for all methods to $\text{2k}$ and compare the results on mAP / mAP50 score.}
\label{tab:comparison}
\renewcommand{\arraystretch}{1.2}
\setlength{\tabcolsep}{3pt}
\resizebox{\textwidth}{!}{%
\begin{tabular}{lccccccccc}
\toprule
 & \multicolumn{2}{c}{Real Data} & & \multicolumn{6}{c}{mAP / mAP50} \\
\cmidrule(lr){2-3} \cmidrule(lr){5-10}
Method & Image & Label & Prec. & YOLOv5-s & YOLOv5-m & YOLOv5-l & YOLOv11-s & YOLOv11-m & YOLOv11-l \\
\midrule
Pre-trained & \checkmark & \checkmark & FP & 37.4 / 56.8 & 45.4 / 64.1 & 49.0 / 67.3 & 47.0 / 65.0 & 51.5 / 70.0 & 53.4 / 72.5 \\
\midrule
LSQ & \checkmark & \checkmark & \multirow{5}{*}{W8A8} & 32.6 / 51.2 & 38.4 / 57.1 & 40.0 / 58.4 & 44.0 / \underline{60.8} & 47.6 / 64.5 & 48.8 / 65.8 \\
LSQ+ & \checkmark & \checkmark & & 32.2 / 51.0 & 38.0 / 56.9 & 39.4 / 58.5 & 43.8 / 60.7 & 47.8 / 64.7 & 48.5 / 65.3 \\
TSOD & $\times$ & \checkmark & & \underline{35.4} / \underline{54.0} & \textbf{42.9} / \textbf{61.4} & \textbf{45.7} / \textbf{63.8} & \textbf{45.7} / \textbf{61.7} & \textbf{50.1} / \underline{65.6} & \underline{51.5} / \underline{66.9} \\
TSOD$^\dagger$ & $\times$ & $\times$ & & \textbf{35.9} / \textbf{54.7} & 37.1 / 53.7 & 30.4 / 46.0 & 43.8 / 58.2 & 44.7 / 56.9 & 44.6 / 55.7 \\
\rowcolor{gray!15}\textbf{GoodQ} & $\times$ & $\times$ & & 35.0 / 53.6 & \underline{42.1} / \underline{60.1} & \underline{45.6} / \underline{63.4} & \underline{45.3} / \textbf{61.7} & \underline{50.0} / \textbf{66.3} & \textbf{51.8} / \textbf{67.8} \\
\midrule
\addlinespace[0.5ex]
LSQ & \checkmark & \checkmark & \multirow{5}{*}{W6A6} & 29.8 / 47.9 & 36.2 / 54.3 & 38.6 / 56.7 & 41.5 / \underline{58.3} & 45.0 / \underline{61.9} & 45.8 / 62.5 \\
LSQ+ & \checkmark & \checkmark & & 29.0 / 47.6 & 35.6 / 53.8 & 38.0 / \underline{57.2} & 41.6 / 58.2 & 44.8 / 61.7 & 45.9 / \underline{62.8} \\
TSOD & $\times$ & \checkmark & & \underline{32.3} / \textbf{50.6} & \textbf{40.4} / \textbf{58.4} & \underline{43.5} / \textbf{61.4} & \underline{42.4} / 57.9 & \underline{45.6} / 60.7 & \underline{46.4} / 60.6 \\
TSOD$^\dagger$ & $\times$ & $\times$ & & 32.0 / \underline{50.0} & 32.7 / 48.1 & 28.8 / 43.1 & 38.1 / 49.9 & 34.5 / 44.2 & 34.1 / 41.7 \\
\rowcolor{gray!15}\textbf{GoodQ} & $\times$ & $\times$ & & \textbf{32.5} / \textbf{50.6} & \underline{40.0} / \underline{58.0} & \textbf{43.6} / \textbf{61.4} & \textbf{43.3} / \textbf{59.4} & \textbf{47.7} / \textbf{63.9} & \textbf{49.3} / \textbf{65.0} \\
\midrule
\addlinespace[0.5ex]
LSQ & \checkmark & \checkmark & \multirow{5}{*}{W4A4} & \underline{19.2} / \underline{34.4} & 27.3 / 44.1 & 31.0 / 48.2 & 29.2 / \underline{43.7} & \underline{35.3} / \underline{50.4} & \underline{36.3} / \underline{51.9} \\
LSQ+ & \checkmark & \checkmark & & 19.0 / 34.0 & 27.4 / 44.5 & 31.1 / 48.2 & \underline{29.3} / 43.4 & 34.8 / 50.1 & \underline{36.3} / \underline{51.9} \\
TSOD & $\times$ & \checkmark & & 18.4 / 32.3 & \underline{28.7} / \underline{45.0} & \underline{33.5} / \underline{50.3} & 14.1 / 21.4 & 16.7 / 23.7 & 20.1 / 28.6 \\
TSOD$^\dagger$ & $\times$ & $\times$ & & 15.0 / 27.9 & 20.3 / 32.0 & 19.3 / 30.8 & 6.58 / 9.61 & 5.95 / 9.00 & 6.22 / 8.89 \\
\rowcolor{gray!15}\textbf{GoodQ} & $\times$ & $\times$ & & \textbf{21.0} / \textbf{36.1} & \textbf{30.7} / \textbf{47.6} & \textbf{34.8} / \textbf{52.0} & \textbf{32.2} / \textbf{45.9} & \textbf{37.4} / \textbf{51.6} & \textbf{39.4} / \textbf{53.6} \\
\midrule
\addlinespace[0.5ex]
LSQ & \checkmark & \checkmark & \multirow{5}{*}{W3A3} & 8.48 / 17.7 & 17.6 / 31.1 & 20.9 / 35.4 & \underline{11.7} / \underline{19.6} & \underline{19.9} / \underline{31.1} & 19.2 / \underline{29.7} \\
LSQ+ & \checkmark & \checkmark & & \underline{9.44} / \underline{19.6} & \underline{17.9} / \underline{31.3} & \underline{21.5} / \underline{35.9} & 11.6 / 19.1 & 18.3 / 28.6 & \textbf{20.4} / \textbf{31.6} \\
TSOD & $\times$ & \checkmark & & 5.0 / 9.9 & 15.0 / 26.5 & 21.1 / 33.8 & 0.5 / 0.7 & 0.8 / 1.5 & 0.6 / 1.1 \\
TSOD$^\dagger$ & $\times$ & $\times$ & & 2.56 / 5.20 & 6.41 / 11.9 & 6.84 / 12.4 & 0.18 / 0.40 & 0.50 / 0.98 & 0.22 / 0.43 \\
\rowcolor{gray!15}\textbf{GoodQ} & $\times$ & $\times$ & & \textbf{10.2} / \textbf{19.7} & \textbf{20.2} / \textbf{33.0} & \textbf{25.4} / \textbf{39.1} & \textbf{14.2} / \textbf{22.1} & \textbf{21.6} / \textbf{31.8} & \underline{19.5} / 28.7 \\
\bottomrule
\end{tabular}
}
\end{table*}

\subsection{Experimental Setting} 

While GoodQ is applicable regardless of the model architecture, we mainly focus on the YOLO family~\cite{yolo}, a single-stage detection architecture widely deployed on edge devices. More specifically, we conducted experiments using YOLOv5~\cite{yolov5} and YOLOv11~\cite{yolov11} models pre-trained on the MS-COCO 2017 dataset~\cite{coco_data}, following TSOD~\cite{zsqod}. We adopted the LSQ~\cite{lsq} scheme to quantize both weights and activations. To demonstrate the effectiveness of our method, we compared our results with LSQ~\cite{lsq}, LSQ+~\cite{lsq_plus}, and TSOD~\cite{zsqod}. Both LSQ and LSQ+ results are reported on a subset of the MS-COCO dataset, whereas the TSOD results are reported on a noise-optimized dataset synthesized using real labels. All experiments were conducted using a training set of size 2k. For a fair comparison, no data augmentation was applied.

\subsection{Results}

Table~\ref{tab:comparison} presents the evaluation results of YOLOv5 and YOLOv11 on the COCO dataset across various QAT and ZSQ-OD methods. The best-performing results are in \textbf{bold}, and the second-best performance is \underline{underlined}. We additionally report a label-free variant of TSOD, which we denote as TSOD$^{\dagger}$ As shown in the table, GoodQ achieves performance comparable to existing approaches in high-bit regimes (e.g., W8A8 and W6A6). While competing methods suffer severe degradation in extreme low-bit regimes (e.g., W4A4 and W3A3), GoodQ remains substantially more robust, exhibiting only a minimal performance drop. This highlights the advantage of leveraging off-the-shelf generative models for ZSQ-OD over traditional optimization-based approaches such as TSOD. Notably, even without access to the original data, our zero-shot approach markedly outperforms TSOD in these low-bit settings, despite TSOD relying on ground-truth labels during QAT optimization. Overall, these results validate the effectiveness of our framework for ZSQ-OD without requiring any access to the original dataset.

\subsection{Ablation Studies}

We conducted three ablation studies to examine the effect of each component of GoodQ independently. Specifically, we varied 1) the training dataset construction process, 2) the QAT loss design, and 3) the size of the image pool. All experiments were conducted with the YOLOv5-s model.

\subsubsection{Ablation on Constructing Training Dataset}

\begin{table}[t]
\centering
\caption{Ablation studies on (a) training dataset construction process and (b) quantization-aware training loss design on mAP / mAP50 score.}
\label{tab:ablations}
\begin{minipage}[t]{0.48\linewidth}
\centering
\textbf{(a) Dataset Construction} \\[0.2cm] 
\setlength{\tabcolsep}{2pt}
\renewcommand{\arraystretch}{1.2}
\begin{tabular}{lcc}
\hline
\textbf{Generation} & \textbf{W8A8} & \textbf{W4A4} \\
\hline
Diffusion & 33.5 / 52.1 & 16.0 / 28.8 \\
+IDP       & 34.3 / 53.5 & 19.3 / 33.8 \\
+IDAS       & 34.4 / 52.9 & 19.8 / 34.4 \\
+IDP + IDAS     & \textbf{35.0} / \textbf{53.6} & \textbf{21.0} / \textbf{36.1} \\
\hline
\end{tabular}
\end{minipage}\hfill
\begin{minipage}[t]{0.48\linewidth}
\centering
\textbf{(b) QAT loss} \\[0.2cm] 
\setlength{\tabcolsep}{2pt}
\renewcommand{\arraystretch}{1.2}
\begin{tabular}{lcc}
\hline
\textbf{Method} & \textbf{W8A8} & \textbf{W4A4} \\
\hline
Base      & 33.9 / 52.8 & 20.6 / 35.8 \\
+TANR & \textbf{35.0} / \textbf{53.6} & \textbf{21.0} / \textbf{36.1} \\
\hline
\end{tabular}
\end{minipage}
\end{table}

Table~\ref{tab:ablations}-(a) presents an ablation study on the training-dataset construction process for QAT under W8A8 and W4A4. \textsc{Diffusion} in the table denotes the baseline, where images generated using GenQ prompts originally designed for classification~\cite{genq} are used as the training set\footnote{We use the GenQ prompt, ``A photo of a $\{\text{D}\}$ $\{\text{C}\}$,'' where $\{\text{D}\}$ is a CLIP ImageNet template and $\{\text{C}\}$ is a randomly sampled class label.}. As shown in the table, both Information-Dense Prompting (IDP) and Intrinsic Distribution-Aware Selection (IDAS) improve ZSQ performance across all bit-widths. Combining the two provides a further boost. Overall, these results confirm that both IDP and IDAS are effective components that reflect the intrinsic characteristics of Object Detection datasets.

\subsubsection{Ablation on QAT Loss Design}

Table~\ref{tab:ablations}-(b) presents an ablation study on the loss functions used for QAT. \textsc{Base} in the table uses the losses proposed in TSOD, specifically combining $\mathcal{L}_\text{distill}$ with the task-specific loss $\mathcal{L}_\text{detect}$ computed using one-hot labels. As shown in the table, replacing $\mathcal{L}_\text{detect}$ with Teacher-guided Adaptive Noise Reduction (TANR) improves performance across all bit regimes. These results indicate that TANR consistently enhances QAT performance.

\subsubsection{Ablation on the Number of Image Pool}
In this paper, we use an image pool of size 160k generated by a generative model. In this section, we conduct an ablation study to analyze how the image-pool size affects ZSQ-OD performance under different quantization bit settings. Table~\ref{tab:image_pool_size} summarizes the results for various image-pool sizes. As shown in the table, increasing the image-pool size generally improves performance, highlighting the benefit of a richer and more diverse set of synthetic images. Notably, the performance gains are more pronounced in low-bit settings, where data diversity plays a more critical role in preserving accuracy under quantization.

\begin{table}[t]
\centering
\caption{Ablation study on the size of the image pool across different quantization bit-widths compared on mAP / mAP50 score.}
\label{tab:image_pool_size}

\setlength{\tabcolsep}{8pt} 
\renewcommand{\arraystretch}{1.2} 

\begin{tabular}{l c c c c}
\toprule
\multirow{2}{*}{\# Image Pool} & \multicolumn{4}{c}{Bit-width} \\
\cmidrule(lr){2-5}
& W8A8 & W6A6 & W4A4 & W3A3 \\
\midrule
2k          & 34.3 / 53.5 & 31.8 / 50.1 & 19.3 / 33.8 & \phantom{0}7.76 / 15.2 \\
16k         & 34.9 / 53.6 & 32.4 / 50.7 & 20.2 / 35.3 & \phantom{0}9.24 / 17.2 \\
32k         & \textbf{35.2} / \textbf{53.9} & 32.5 / \textbf{50.9} & 21.1 / \textbf{36.7} & \phantom{0}9.41 / 17.5 \\
64k         & 35.1 / 53.8 & \textbf{32.6} / \textbf{50.9} & \textbf{21.3} / 36.5 & \phantom{0}9.20 / 17.0 \\
160k (Ours) & 35.0 / 53.6 & 32.5 / 50.6 & 21.0 / 36.1 &  \textbf{10.20} / \textbf{19.7} \\
\bottomrule
\end{tabular}
\end{table}

\begin{figure*}[t]
    \centering
    \begin{minipage}[c]{0.48\textwidth}
        \centering
        \includegraphics[width=1.0\linewidth]{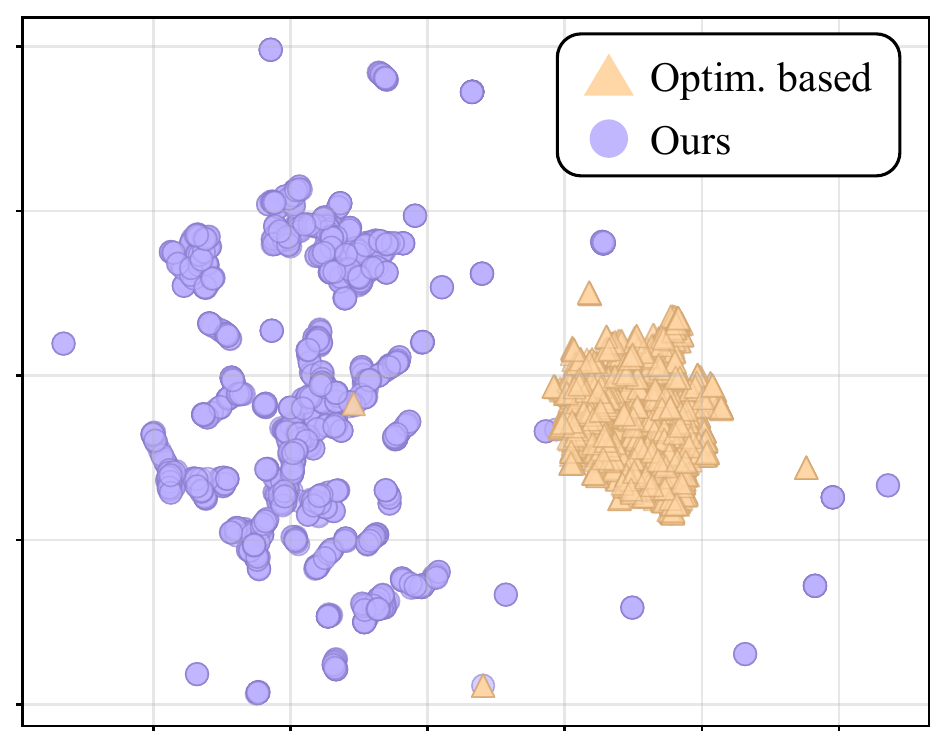}
    \caption{UMAP visualization of CLIP embeddings for the optimization-based synthesized dataset and Ours.}
    \label{fig:diversity_tsne_clip}
    \end{minipage}
    \hfill
    \begin{minipage}[c]{0.48\textwidth}
        \centering
        \captionof{table}{Cosine distance measured on the optimization-based synthesized training dataset and ours across various quantization settings. Higher values indicate greater diversity. Ours show higher diversity across all bit rates.}
        \label{tab:cosine_distance_results}
        \renewcommand{\arraystretch}{1.2}
        \begin{tabular}{lccc}
        \hline
        \textbf{Dataset} & \textbf{FP} & \textbf{W8A8} & \textbf{W4A4} \\
        \hline
        Optim. & 0.539 & 0.535 & 0.133 \\
        \textbf{Ours} & \textbf{0.665} & \textbf{0.665} & \textbf{0.248} \\
        \hline
        \end{tabular}
    \end{minipage}
\end{figure*}

\section{Analysis}

In this section, we provide analyses to answer why GoodQ is effective. We first analyze GoodQ from the perspective of dataset generation and then examine it in the context of the QAT process. All analyses are performed using the YOLOv5-s.

\subsection{Analysis on Dataset Generation Perspective}
\label{sec:analysis_dataset}

We first analyze the dataset generation in GoodQ by examining two aspects: 1) the importance of data diversity in the low-bit regime and 2) the need to align the class distribution of the constructed training dataset. For the first analysis, we compare the noise-optimization-based synthetic dataset with the dataset constructed using a generative model. For the second analysis, to evaluate the importance of distribution alignment, we incorporate different image selection methods and measure QAT performance on the resulting training sets.

\subsubsection{Finding 1. Diversity Matters in Low-bit}

As demonstrated in Table~\ref{tab:comparison}, performing ZSQ-OD with a diffusion-generated training dataset yields superior performance in extreme low-bit regimes (e.g., W4A4 and W3A3) compared to noise-optimization-based methods such as TSOD. To understand the source of this advantage, we analyze the feature representations of the two datasets. Figure~\ref{fig:diversity_tsne_clip} shows a UMAP~\cite{umap} visualization of their CLIP~\cite{clip} embeddings, indicating that the diffusion-generated dataset is substantially more dispersed. This observation is quantitatively supported by Table~\ref{tab:cosine_distance_results}, which reports the average pairwise cosine distance\footnote{Let $(x_i)_{i=1}^{N}$ denote embeddings extracted from $N$ images. We compute the cosine distance for each image pair $(i,j)$ as $d_{\mathrm{cos}}(x_i,x_j)=1-\cos(x_i,x_j)$. The final number reported in the table was calculated by averaging over every pairwise cosine distance $d_{\mathrm{cos}}(x_i,x_j)$ (typically $i<j$).} computed from embeddings extracted by both full-precision and quantized models. These results confirm the superior diversity of the diffusion-generated images. Such enhanced diversity appears particularly beneficial under severe quantization constraints. By covering a broader region of the feature space, the synthetic dataset can help the heavily quantized student model capture the knowledge of the full-precision teacher more effectively~\cite{gooddistill}. Overall, our analysis suggests that maximizing training-data diversity is an important factor for robust low-bit ZSQ-OD performance.

\begin{table}[t]
\centering
\caption{Comparison of the class-wise BBox distribution between the real COCO training dataset and the distribution resulting from diverse selection methods.}
\label{tab:distribution_comparison}
\setlength{\tabcolsep}{8pt}
\renewcommand{\arraystretch}{1.2}
\begin{tabular}{lcc|ccc}
\hline
\textbf{Selection} & \textbf{W8A8} & \textbf{W4A4} & \textbf{\#BBox} & \textbf{KL} & \textbf{L1}  \\
\hline
Random &  34.3 / 53.5 &  19.3 / 33.8 & 7917 & 0.48& 0.77  \\
Many BBox &  33.5 / 51.7 &  17.5 / 30.7 & \textbf{72786} & 1.25 & 1.14  \\
\textbf{IDAS} & \textbf{35.0}/ \textbf{53.6} &  \textbf{21.0} / \textbf{36.1} & 25333 & \textbf{0.2616} & \textbf{0.6116} \\
\hline
\end{tabular}
\end{table}

\subsubsection{Finding 2. Class Distribution Matters}

As illustrated in Table~\ref{tab:ablations}-(a), estimating and curating a dataset that matches the original skewed class-wise distribution improves QAT performance in the low-bit regime. To further investigate the importance of aligning generated data with the class-wise bounding-box distribution, we apply three selection algorithms after constructing an image pool using Information-Dense Prompting (IDP). Table~\ref{tab:distribution_comparison} provides a comprehensive comparison of performance and statistics for random selection (\textsc{Random}), naive selection based on the number of bounding boxes (\textsc{Many BBox}), and IDAS. The superior performance of IDAS over \textsc{Random} suggests that accurate distribution matching directly improves QAT performance. Notably, the comparison with \textsc{Many BBox} indicates that the gains are not merely a byproduct of selecting images with more bounding boxes. As shown in the table, this naive strategy degrades performance.

These results suggest that explicitly reflecting the intrinsic characteristics of the original dataset during training-data construction is crucial for successful ZSQ-OD. As shown in the same table, the dataset selected by IDAS matches the original distribution more closely than \textsc{Random}, as evidenced by lower KL divergence and L1 distance relative to the original class-wise distribution\footnote{The class-wise BBox distribution of the original OD dataset is computed as the ratio of BBoxes belonging to each class to the total number of BBoxes in the training set.}. In contrast, \textsc{Many BBox} selection deviates further from the original distribution. Overall, these findings confirm that the QAT training dataset should reflect the true class-wise distribution.

\subsection{Analysis on QAT Perspective}

Next, we analyze the QAT process in GoodQ by examining the importance of mitigating pseudo-label noise before QAT. Specifically, we compare the impact of Teacher-guided Adaptive Noise Reduction (TANR) across two distinct training datasets: one generated by a diffusion model and the other constructed via optimization-based synthesis.

\begin{figure*}[t]
    \centering
    \begin{minipage}[c]{0.48\linewidth}
        \centering
        \captionof{table}{Comparison of the quantization performance before and after applying TANR to both ours dataset and optimization-based images. \textsc{Ours} indicates the resulting set gathered by applying IDAS to the diffusion image pool.The results in the table report performance in terms of mAP/mAP50.}
        \label{tab:qfocal_sft_relation_2}

        \resizebox{\linewidth}{!}{
            \renewcommand{\arraystretch}{1.3}
            \begin{tabular}{c l c c}
            \hline
            \textbf{Precision} & \textbf{Method} & \textbf{Ours} & \textbf{Optim.} \\
            \hline
            FP & - & \multicolumn{2}{c}{37.4 / 56.8} \\
            \hline
            \multirow{2}{*}{W8A8}
            & Base  & 33.9 / 52.8 & 35.4 / 54.0 \\
            & +TANR & 35.0 / 53.6 & 35.7 / 54.4 \\
            \hline
            \multirow{2}{*}{W4A4}
            & Base  & 20.6 / 35.8 & 18.4 / 32.3 \\
            & +TANR & 21.0 / 36.1 & 18.3 / 32.1 \\
            \hline
            \end{tabular}
        } 
    \end{minipage}%
    \hfill
    \begin{minipage}[c]{0.5\linewidth}
        \centering
        \includegraphics[width=1.0\linewidth]{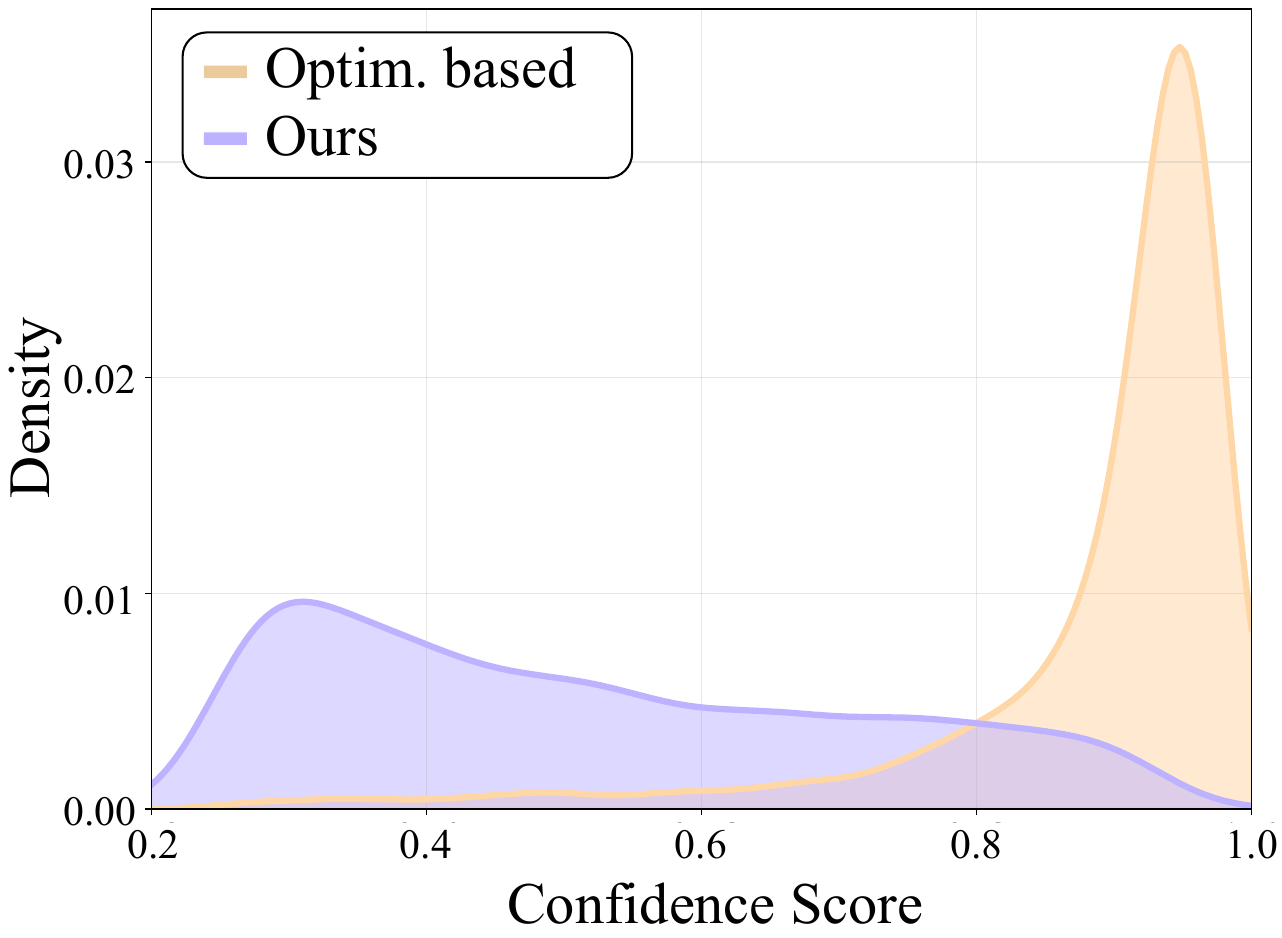}
        \captionof{figure}{Classification confidence distributions of localized objects under different generation configurations are shown. For improved visualization, the overall distributions are smoothed.}
        \label{fig:confidence_distribution}
    \end{minipage}
\end{figure*}

\subsubsection{Finding 3. Teacher Noise Matters} 

Table~\ref{tab:qfocal_sft_relation_2} presents a performance comparison when TANR is applied to training datasets constructed via optimization-based synthesis versus our diffusion-based generation. As shown in the table, TANR consistently improves performance on the diffusion-generated dataset, whereas its gains on the optimization-based baseline are marginal. The reason for this discrepancy is illustrated in Figure~\ref{fig:confidence_distribution}, showing that diffusion-generated images yield lower confidence scores\footnote{The confidence score is computed as the product of the objectness score and the class confidence: $p_{\text{conf}} = \sigma(\hat{y}_o) \cdot \max(\sigma(\mathbf{\hat{y}}_c))$.} than images synthesized by optimization-based methods. These results indicate that TANR is effective for addressing the challenges that arise when using generative models for ZSQ-OD.
\section{Conclusion}

In this paper, we introduce GoodQ (\textbf{G}enerative \textbf{o}ff-the-shelf models for \textbf{o}bject \textbf{d}etector \textbf{Q}uantization), a novel framework designed to improve Zero-Shot Quantization for Object Detection (ZSQ-OD) by leveraging off-the-shelf generative models in a task-specific manner. For Object Detection, GoodQ accounts for challenges such as the need for high-density information within a single frame and highly skewed class-wise distributions. Moreover, generative models introduce additional complexity because pseudo-labels are inherently noisy. To address these challenges, GoodQ incorporates three synergistic components: Information-Dense Prompting (IDP), Intrinsic Distribution-Aware Selection (IDAS), and Teacher-guided Adaptive Noise Reduction (TANR). Extensive experiments across detection models show that GoodQ substantially mitigates performance degradation in challenging low-bit regimes---where optimization-based synthesis methods struggle---while maintaining competitive performance in high-bit regimes. Finally, we provide in-depth analyses across multiple dimensions, offering practical insights for successful ZSQ-OD.


\section*{Acknowledgements}
This work was supported by the Korean Government through the grants from IITP (RS-2021-II211343, RS-2025-25442338, 26-InnoCORE-01).


%
%
\bibliographystyle{splncs04}
\bibliography{main}

@String(CVPR  = {IEEE Conf. Comput. Vis. Pattern Recog.})

@String(ICCV  = {Int. Conf. Comput. Vis.})

@String(AAAI  = {AAAI})

@String(CVPR  = {CVPR})

@String(ICCV  = {ICCV})

@InProceedings{lsq_plus,
author = {Bhalgat, Yash and Lee, Jinwon and Nagel, Markus and Blankevoort, Tijmen and Kwak, Nojun},
title = {LSQ+: Improving Low-Bit Quantization Through Learnable Offsets and Better Initialization},
booktitle = {Proceedings of the IEEE/CVF Conference on Computer Vision and Pattern Recognition (CVPR) Workshops},
month = {June},
year = {2020}
}

@article{qkd,
  title={Qkd: Quantization-aware knowledge distillation},
  author={Kim, Jangho and Bhalgat, Yash and Lee, Jinwon and Patel, Chirag and Kwak, Nojun},
  journal={arXiv preprint arXiv:1911.12491},
  year={2019}
}

@article{lsq,
  title={Learned step size quantization},
  author={Esser, Steven K and McKinstry, Jeffrey L and Bablani, Deepika and Appuswamy, Rathinakumar and Modha, Dharmendra S},
  journal={arXiv preprint arXiv:1902.08153},
  year={2019}
}

@inproceedings{genq,
  title={Genq: Quantization in low data regimes with generative synthetic data},
  author={Li, Yuhang and Kim, Youngeun and Lee, Donghyun and Kundu, Souvik and Panda, Priyadarshini},
  booktitle={European Conference on Computer Vision},
  pages={216--235},
  year={2024},
  organization={Springer}
}

@inproceedings{synq,
  title={Synq: Accurate zero-shot quantization by synthesis-aware fine-tuning},
  author={Kim, Minjun and Kim, Jongjin and Kang, U},
  booktitle={The Thirteenth International Conference on Learning Representations},
  year={2025}
}

@inproceedings{zeroq,
  title={Zeroq: A novel zero shot quantization framework},
  author={Cai, Yaohui and Yao, Zhewei and Dong, Zhen and Gholami, Amir and Mahoney, Michael W and Keutzer, Kurt},
  booktitle={Proceedings of the IEEE/CVF conference on computer vision and pattern recognition},
  pages={13169--13178},
  year={2020}
}

@InProceedings{mixup_genq,
    author    = {Park, Jiwoong and Lee, Chaeun and Choi, Yongseok and Park, Sein and Hong, Deokki and Choi, Jungwook},
    title     = {Enhancing Generalization in Data-free Quantization via Mixup-class Prompting},
    booktitle = {Proceedings of the IEEE/CVF International Conference on Computer Vision (ICCV) Workshops},
    month     = {October},
    year      = {2025},
    pages     = {4002-4011}
}

@inproceedings{ris,
  title={Robustness-guided image synthesis for data-free quantization},
  author={Bai, Jianhong and Yang, Yuchen and Chu, Huanpeng and Wang, Hualiang and Liu, Zuozhu and Chen, Ruizhe and He, Xiaoxuan and Mu, Lianrui and Cai, Chengfei and Hu, Haoji},
  booktitle={Proceedings of the AAAI Conference on Artificial Intelligence},
  volume={38},
  pages={10971--10979},
  year={2024}
}

@inproceedings{gdfq,
  title={Generative low-bitwidth data free quantization},
  author={Xu, Shoukai and Li, Haokun and Zhuang, Bohan and Liu, Jing and Cao, Jiezhang and Liang, Chuangrun and Tan, Mingkui},
  booktitle={European conference on computer vision},
  pages={1--17},
  year={2020},
  organization={Springer}
}

@article{diffusion,
  title={Denoising diffusion probabilistic models},
  author={Ho, Jonathan and Jain, Ajay and Abbeel, Pieter},
  journal={Advances in neural information processing systems},
  volume={33},
  pages={6840--6851},
  year={2020}
}

@inproceedings{zsqod,
  title={Task-Specific Zero-shot Quantization-Aware Training for Object Detection},
  author={Li, Changhao and Chen, Xinrui and Wang, Ji and Zhao, Kang and Chen, Jianfei},
  booktitle={Proceedings of the IEEE/CVF International Conference on Computer Vision},
  pages={22868--22878},
  year={2025}
}

@inproceedings{clip,
  title={Learning transferable visual models from natural language supervision},
  author={Radford, Alec and Kim, Jong Wook and Hallacy, Chris and Ramesh, Aditya and Goh, Gabriel and Agarwal, Sandhini and Sastry, Girish and Askell, Amanda and Mishkin, Pamela and Clark, Jack and others},
  booktitle={International conference on machine learning},
  pages={8748--8763},
  year={2021},
  organization={PmLR}
}

@inproceedings{coco_data,
  title={Microsoft coco: Common objects in context},
  author={Lin, Tsung-Yi and Maire, Michael and Belongie, Serge and Hays, James and Perona, Pietro and Ramanan, Deva and Doll{\'a}r, Piotr and Zitnick, C Lawrence},
  booktitle={European conference on computer vision},
  pages={740--755},
  year={2014},
  organization={Springer}
}

@inproceedings{yolo,
  title={You only look once: Unified, real-time object detection},
  author={Redmon, Joseph and Divvala, Santosh and Girshick, Ross and Farhadi, Ali},
  booktitle={Proceedings of the IEEE conference on computer vision and pattern recognition},
  pages={779--788},
  year={2016}
}

@article{yolov5,
  title={What is YOLOv5: A deep look into the internal features of the popular object detector},
  author={Khanam, Rahima and Hussain, Muhammad},
  journal={arXiv preprint arXiv:2407.20892},
  year={2024}
}

@article{yolov11,
  title={Yolov11: An overview of the key architectural enhancements},
  author={Khanam, Rahima and Hussain, Muhammad},
  journal={arXiv preprint arXiv:2410.17725},
  year={2024}
}

@inproceedings{fcn,
  title={Fully quantized network for object detection},
  author={Li, Rundong and Wang, Yan and Liang, Feng and Qin, Hongwei and Yan, Junjie and Fan, Rui},
  booktitle={Proceedings of the IEEE/CVF conference on computer vision and pattern recognition},
  pages={2810--2819},
  year={2019}
}

@inproceedings{mimiq,
  title={Mimiq: Low-bit data-free quantization of vision transformers with encouraging inter-head attention similarity},
  author={Choi, Kanghyun and Lee, Hyeyoon and Kwon, Dain and Park, SunJong and Kim, Kyuyeun and Park, Noseong and Choi, Jonghyun and Lee, Jinho},
  booktitle={Proceedings of the AAAI Conference on Artificial Intelligence},
  volume={39},
  pages={16037--16045},
  year={2025}
}

@inproceedings{clampvit,
  title={Clamp-vit: Contrastive data-free learning for adaptive post-training quantization of vits},
  author={Ramachandran, Akshat and Kundu, Souvik and Krishna, Tushar},
  booktitle={European Conference on Computer Vision},
  pages={307--325},
  year={2024},
  organization={Springer}
}

@inproceedings{hast,
  title={Hard sample matters a lot in zero-shot quantization},
  author={Li, Huantong and Wu, Xiangmiao and Lv, Fanbing and Liao, Daihai and Li, Thomas H and Zhang, Yonggang and Han, Bo and Tan, Mingkui},
  booktitle={Proceedings of the IEEE/CVF conference on Computer Vision and Pattern Recognition},
  pages={24417--24426},
  year={2023}
}

@article{qimera,
  title={Qimera: Data-free quantization with synthetic boundary supporting samples},
  author={Choi, Kanghyun and Hong, Deokki and Park, Noseong and Kim, Youngsok and Lee, Jinho},
  journal={Advances in Neural Information Processing Systems},
  volume={34},
  pages={14835--14847},
  year={2021}
}

@inproceedings{intraq,
  title={Intraq: Learning synthetic images with intra-class heterogeneity for zero-shot network quantization},
  author={Zhong, Yunshan and Lin, Mingbao and Nan, Gongrui and Liu, Jianzhuang and Zhang, Baochang and Tian, Yonghong and Ji, Rongrong},
  booktitle={Proceedings of the IEEE/CVF Conference on Computer Vision and Pattern Recognition},
  pages={12339--12348},
  year={2022}
}

@inproceedings{enhancingDFQ,
  title={Enhancing diversity for data-free quantization},
  author={Zhao, Kai and Zhuang, Zhihao and Zhang, Miao and Guo, Chenjuan and Shu, Yang and Yang, Bin},
  booktitle={Proceedings of the Computer Vision and Pattern Recognition Conference},
  pages={20969--20978},
  year={2025}
}

@article{genvit,
  title={Joint Post-Training Quantization of Vision Transformers with Learned Prompt-Guided Data Generation},
  author={Li, Shile and Karmann, Markus and Urfalioglu, Onay},
  journal={arXiv preprint arXiv:2602.18861},
  year={2026}
}

@article{qfocal,
  title={Generalized focal loss: Learning qualified and distributed bounding boxes for dense object detection},
  author={Li, Xiang and Wang, Wenhai and Wu, Lijun and Chen, Shuo and Hu, Xiaolin and Li, Jun and Tang, Jinhui and Yang, Jian},
  journal={Advances in neural information processing systems},
  volume={33},
  pages={21002--21012},
  year={2020}
}

@article{umap,
  title={Umap: Uniform manifold approximation and projection for dimension reduction},
  author={McInnes, Leland and Healy, John and Melville, James},
  journal={arXiv preprint arXiv:1802.03426},
  year={2018}
}

@inproceedings{gooddistill,
  title={What Makes a Good Dataset for Knowledge Distillation?},
  author={Frank, Logan and Davis, Jim},
  booktitle={Proceedings of the Computer Vision and Pattern Recognition Conference},
  pages={23755--23764},
  year={2025}
}

@inproceedings{he2017mask,
  title={Mask r-cnn},
  author={He, Kaiming and Gkioxari, Georgia and Doll{\'a}r, Piotr and Girshick, Ross},
  booktitle={Proceedings of the IEEE international conference on computer vision},
  pages={2961--2969},
  year={2017}
}

@inproceedings{liu2021swin,
  title={Swin transformer: Hierarchical vision transformer using shifted windows},
  author={Liu, Ze and Lin, Yutong and Cao, Yue and Hu, Han and Wei, Yixuan and Zhang, Zheng and Lin, Stephen and Guo, Baining},
  booktitle={Proceedings of the IEEE/CVF international conference on computer vision},
  pages={10012--10022},
  year={2021}
}

@misc{Homeobjects,
    author = {Jocher, Glenn and Rizwan, Muhammad},
    license = {AGPL-3.0},
    month = {May},
    title = {Ultralytics Datasets: HomeObjects-3K Detection Dataset},
    url = {https://docs.ultralytics.com/datasets/detect/homeobjects-3k/},
    version = {1.0.0},
    year = {2025}
}

@misc{VOC,
      title={The PASCAL Visual Object Classes (VOC) Challenge},
      author={Mark Everingham and Luc Van Gool and Christopher K. I. Williams and John Winn and Andrew Zisserman},
      year={2010},
      eprint={0909.5206},
      archivePrefix={arXiv},
      primaryClass={cs.CV}
}

\clearpage
\appendix
\section{Details on Prompts}

\subsection{Prompt Template for Information-Dense Prompting}

\begin{figure*}[h] 
    \centering
    \includegraphics[width=1.0\linewidth]{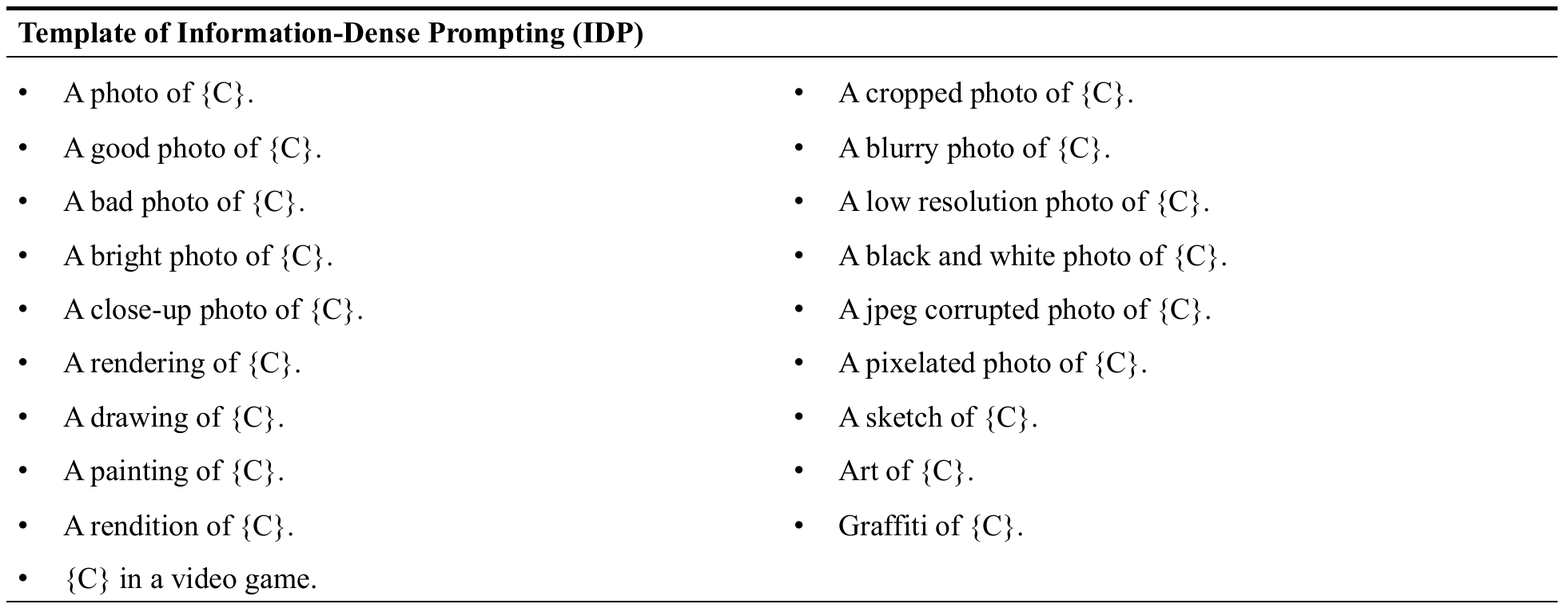} 
    
    \caption{Template of Information-Dense Prompting (IDP) used in the image generation.}
    \label{fig:idp_prompt}
\end{figure*}

\km{In this paper, we utilized Information-Dense Prompting (IDP) to generate images suitable for object detection. Figure~\ref{fig:idp_prompt} illustrates the prompt template used for image generation. Additionally, for the Quantity, \{Q\}, we employed a total of six terms: \textsc{multiple}, \textsc{many}, \textsc{several}, \textsc{a few}, \textsc{some}, and \textsc{numerous}. Furthermore, when generating images via IDP, we randomly selected two distinct class names for each generation iteration. For example, the final prompt can be ``A photo of many cars and a few person''.}

\begin{figure*}[t] 
    \centering
    \includegraphics[width=1.0\linewidth]{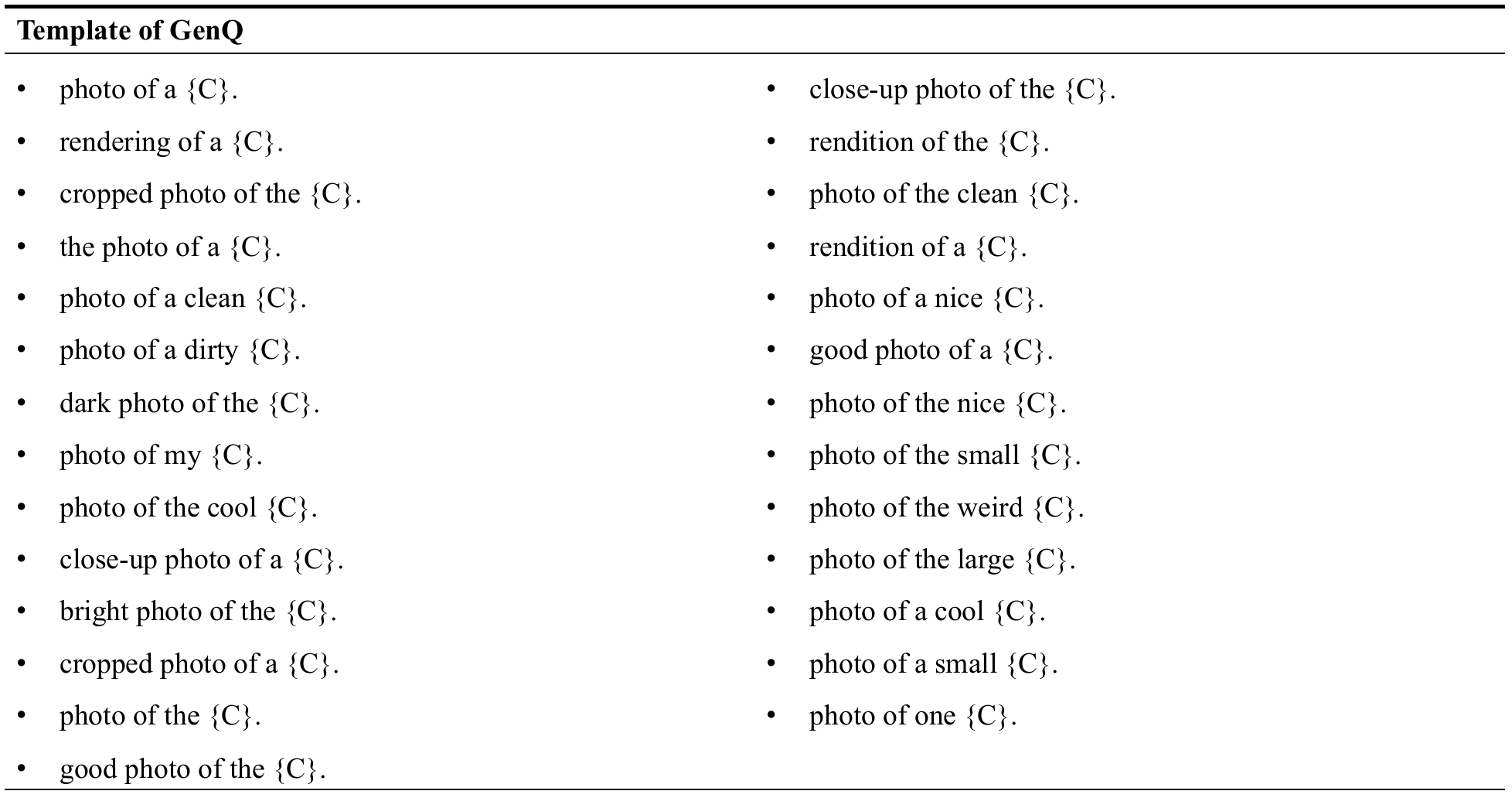} 
    
    \caption{Template of GenQ used in the image generation process.}
    \label{fig:genq}
\end{figure*}

\km{For the experiments in Figure~\ref{fig:teaser} and Table~\ref{tab:ablations}, we utilized the prompts introduced in a Zero-Shot Quantization paper for image classification. Figure~\ref{fig:genq} illustrates the prompt template employed in these experiments, which is identical to the one used in GenQ~\cite{genq}. Consistent with the IDP approach, we randomly selected the classes when generating images using this prompt.}

\subsection{Prompt Template for GenQ}

\km{For the experiments in Figure~\ref{fig:teaser} and Table~\ref{tab:ablations}, we utilized the prompts introduced in a Zero-Shot Quantization paper for image classification. Figure~\ref{fig:genq} illustrates the prompt template employed in these experiments, which is identical to the one used in GenQ~\cite{genq}. Consistent with the IDP approach, we randomly selected the classes when generating images using this prompt.}

\section{Details on Experimental Settings}

\subsection{Detailed Experimental Setting of GoodQ}

\subsubsection{Generation}
\km{To construct the image pool, we first generate images using Stable Diffusion v1.5, and subsequently perform pseudo-labeling with one-hot labels to facilitate Intrinsic Distribution-Aware Selection (IDAS). The pseudo-labels are given using a detection model. Detailed configurations are provided in Table~\ref{tab:exp_config_gen}.}

\begin{table}[htbp]
\centering
\caption{Experimental configurations in the image generation stage.}
\label{tab:exp_config_gen}
\begin{tabular}{lcc}
\toprule
\textbf{Hyperparameters} & \textbf{YOLOv5} & \textbf{YOLOv11} \\
\midrule
Guidance scale & 7.5 & 7.5 \\
Confidence threshold & 0.25 & 0.25 \\
IOU threshold & 0.45 & 0.70 \\
Inference steps & 50 & 50 \\
Image resolution & 640 $\times$ 640 & 640 $\times$ 640 \\
\bottomrule
\end{tabular}
\end{table}

\subsubsection{QAT} \km{In this section, we detail the experimental configurations employed for GoodQ. Table~\ref{tab:exp_config} summarizes the specific hyperparameters used for training the YOLOv5 and YOLOv11 architectures. Both models were trained for 100 epochs using the Adam optimizer. This overall configuration strictly adheres to the settings established in TSOD~\cite{zsqod}. Additionally, the reproduced performance metrics presented in Table~\ref{tab:comparison} were obtained by executing the original code-bases with their default settings to ensure a fair and consistent comparison.}

\begin{table}[t]
\centering
\caption{Experimental configurations in the QAT process for YOLOv5 and YOLOv11.}
\label{tab:exp_config}
\begin{tabular}{lcc}
\toprule
\textbf{Hyperparameters} & \textbf{YOLOv5} & \textbf{YOLOv11} \\
\midrule
Batch Size & 32 (s) / 16 (m) / 8 (l) & 16 \\
Optimizer & Adam & Adam \\
Learning Rate & $0.01$ & $0.00001$ \\
Epochs & 100 & 100 \\
QFocal $\gamma$ & $1.5$ & $2.0$ \\
$\mathcal{L}_\text{KD}$ (Temp. / Coeff.) & 4 / $0.1$ & 4 / $0.1$ \\
$\mathcal{L}_\text{MSE}$ (Coeff.) & $1.0$ & $1.0$ \\
$\mathcal{L}_\text{detect}$ (Coeff.) & $0.04$ & $0.01$ \\
\bottomrule
\end{tabular}
\end{table}

\subsection{Detailed Explanation of Loss Function}

\km{For effective Quantization-Aware Training, TSOD~\cite{zsqod} utilizes $\mathcal{L}_{\text{distill}}$ and $\mathcal{L}_{\text{detect}}$ during training. This section introduces the composition of $\mathcal{L}_{\text{distill}}$. Unlike $\mathcal{L}_{\text{detect}}$, $\mathcal{L}_{\text{distill}}$ employs a task-agnostic approach to narrow the knowledge gap between the full-precision and quantized models. Specifically, $\mathcal{L}_{\text{distill}}$ consists of two components: 1) prediction-matching distillation ($\mathcal{L}_{\text{KD}}$) and 2) feature-level distillation ($\mathcal{L}_{\text{feat}}$). }\km{For $\mathcal{L}_{\text{KD}}$, predictions from all detection heads (e.g., three heads with different strides in the case of YOLOv5) of both the full-precision and quantized detection models are first concatenated and then flattened. The loss is subsequently calculated using the following equation:}

\begin{align}\mathcal{L}_\text{KD} = \frac{\tau^2}{N} \sum_{i=1}^{N} \text{KL}(z^F(\hat{x}_i; \theta), z^Q(\hat{x}_i; \theta')),
\end{align}

\noindent \km{where $\tau$ represents the temperature parameter, and $z^F$ and $z^Q$ denote the full-precision ($\theta$) and quantized models ($\theta'$), respectively.}

\km{Furthermore, $\mathcal{L}_{\text{feat}}$ is applied to the intermediate features of the detection models, and the training proceeds using the following loss function:}

\begin{align}\mathcal{L}_{feat} = \frac{1}{NL} \sum_{i=1}^{N} \sum_{l=1}^{L} | f_l^F(\hat{x}_i; \theta) - f_l^Q(\hat{x}_i; \theta') |_2^2,
\end{align}

\noindent \km{where $f_l^F$ and $f_l^Q$ represent the features of the full-precision ($\theta$) and quantized models ($\theta'$) at the intermediate layer $l$, respectively.}

\section{Further Ablations and Examples}

\subsection{GoodQ on Different Quantization Methods}

\km{We conducted Quantization-Aware Training (QAT) based on LSQ~\cite{lsq}, following the same approach as TSOD~\cite{zsqod}. However, GoodQ is not limited to LSQ and can be readily applied to other quantization methods. Table~\ref{tab:comparison_abl} presents the results of applying GoodQ to LSQ+ (GoodQ+)~\cite{lsq_plus} and TSOD to LSQ+ (TSOD+).}

\begin{table*}[t]
\centering
\caption{Comparison with real data Quantization Aware Training (QAT) on YOLOv5/YOLOv11 on MS-COCO validation set. We fix the size of the calibration set for all methods to $\text{2k}$ and compare the results on mAP / mAP50 score.}
\label{tab:comparison_abl}
\renewcommand{\arraystretch}{1.2}
\setlength{\tabcolsep}{3pt}
\resizebox{\textwidth}{!}{%
\begin{tabular}{lccccccccc}
\toprule
 & \multicolumn{2}{c}{Real Data} & & \multicolumn{6}{c}{mAP / mAP50} \\
\cmidrule(lr){2-3} \cmidrule(lr){5-10}
Method & Image & Label & Prec. & YOLOv5-s & YOLOv5-m & YOLOv5-l & YOLOv11-s & YOLOv11-m & YOLOv11-l \\
\midrule
Pre-trained & \checkmark & \checkmark & FP & 37.4 / 56.8 & 45.4 / 64.1 & 49.0 / 67.3 & 47.0 / 65.0 & 51.5 / 70.0 & 53.4 / 72.5 \\
\midrule
LSQ & \checkmark & \checkmark & \multirow{4}{*}{W8A8} & 32.6 / 51.2 & 38.4 / 57.1 & 40.0 / 58.4 & 44.0 / 60.8 & 47.6 / 64.5 & 48.8 / 65.8 \\
LSQ+ & \checkmark & \checkmark & & 32.2 / 51.0 & 38.0 / 56.9 & 39.4 / 58.5 & 43.8 / 60.7 & 47.8 / 64.7 & 48.5 / 65.3 \\
TSOD+ & $\times$ & \checkmark & & \textbf{35.4} / \textbf{54.3} & \textbf{42.8} / \textbf{61.4} & \textbf{45.8} / \textbf{63.7} & \textbf{45.7} / \textbf{61.8} & \textbf{50.0} / \underline{65.9} & \textbf{51.7} / \underline{66.9} \\
\rowcolor{gray!15}GoodQ+ & $\times$ & $\times$ & & \underline{35.2} / \underline{53.8} & \underline{42.2} / \underline{60.1} & \underline{45.3} / \underline{63.3} & \underline{45.3} / \textbf{61.8} & \underline{49.9} / \textbf{66.2} & \textbf{51.7} / \textbf{67.9} \\
\midrule
\addlinespace[0.5ex]
LSQ & \checkmark & \checkmark & \multirow{4}{*}{W6A6} & 29.8 / 47.9 & 36.2 / 54.3 & 38.6 / 56.7 & 41.5 / \underline{58.3} & 45.0 / \underline{61.9} & 45.8 / 62.5 \\
LSQ+ & \checkmark & \checkmark & & 29.0 / 47.6 & 35.6 / 53.8 & 38.0 / 57.2 & 41.6 / 58.2 & 44.8 / 61.7 & 45.9 / \underline{62.8} \\
TSOD+ & $\times$ & \checkmark & & \textbf{32.6} / \textbf{50.9} & \textbf{40.3} / \textbf{58.2} & \textbf{43.5} / \textbf{61.5} & \underline{42.4} / 57.8 & \underline{45.7} / 60.7 & \underline{46.1} / 60.2 \\
\rowcolor{gray!15}GoodQ+ & $\times$ & $\times$ & & \underline{32.3} / \underline{50.4} & \underline{39.7} / \underline{58.0} & \textbf{43.5} / \underline{61.2} & \textbf{43.5} / \textbf{59.6} & \textbf{47.8} / \textbf{63.9} & \textbf{49.3} / \textbf{65.0} \\
\midrule
\addlinespace[0.5ex]
LSQ & \checkmark & \checkmark & \multirow{4}{*}{W4A4} & \underline{19.2} / \underline{34.4} & 27.3 / 44.1 & 31.0 / 48.2 & 29.2 / \underline{43.7} & \underline{35.3} / \underline{50.4} & \underline{36.3} / \underline{51.9} \\
LSQ+ & \checkmark & \checkmark & & 19.0 / 34.0 & 27.4 / 44.5 & 31.1 / 48.2 & \underline{29.3} / 43.4 & 34.8 / 50.1 & \underline{36.3} / \underline{51.9} \\
TSOD+ & $\times$ & \checkmark & & 18.4 / 32.5 & \underline{28.6} / \underline{44.7} & \underline{33.8} / \underline{50.3} & 16.2 / 24.2 & 15.1 / 21.9 & 21.5 / 30.4 \\
\rowcolor{gray!15}GoodQ+ & $\times$ & $\times$ & & \textbf{21.1} / \textbf{36.3} & \textbf{30.8} / \textbf{47.6} & \textbf{35.0} / \textbf{52.2} & \textbf{31.4} / \textbf{44.9} & \textbf{37.9} / \textbf{51.9} & \textbf{39.8} / \textbf{54.3} \\
\midrule
\addlinespace[0.5ex]
LSQ & \checkmark & \checkmark & \multirow{4}{*}{W3A3} & 8.48 / 17.7 & 17.6 / 31.1 & 20.9 / 35.4 & \underline{11.7} / \underline{19.6} & \underline{19.9} / \underline{31.1} & 19.2 / \underline{29.7} \\
LSQ+ & \checkmark & \checkmark & & \underline{9.44} / \underline{19.6} & \underline{17.9} / \underline{31.3} & \underline{21.5} / \underline{35.9} & 11.6 / 19.1 & 18.3 / 28.6 & \textbf{20.4} / \textbf{31.6} \\
TSOD+ & $\times$ & \checkmark & & 5.18 / \textbf{21.1} & 15.3 / 26.6 & 21.3 / 34.6 & 0.58 / 1.15 & 0.98 / 1.8 & 0.48 / 0.97 \\
\rowcolor{gray!15}GoodQ+ & $\times$ & $\times$ & & \textbf{9.52} / 18.2 & \textbf{20.1} / \textbf{32.9} & \textbf{25.3} / \textbf{39.4} & \textbf{14.3} / \textbf{22.2} & \textbf{21.7} / \textbf{31.7} & \underline{19.4} / 28.6 \\
\bottomrule
\end{tabular}
}
\end{table*}

\begin{table}[t]
\centering
\caption{Performance comparison across different quantization bit-widths on mask R-CNN architecture with Swin-T (tiny) backbone.}
\label{tab:transformer}
\setlength{\tabcolsep}{6.5pt} 
\renewcommand{\arraystretch}{1.1} 
\begin{tabular}{lcc cccc}
\toprule
& \multicolumn{2}{c}{Real Data} & \multicolumn{4}{c}{mAP / mAP50} \\
\cmidrule(lr){2-3} \cmidrule(lr){4-7}
\textbf{Method} & Image & Label & \textbf{W8A8} & \textbf{W6A6} & \textbf{W4A4} & \textbf{W3A3} \\
\midrule
LSQ  & $\checkmark$ & $\checkmark$ & 44.9 / 66.8 & 42.6 / 64.2 & 34.5 / 54.7 & 23.8 / 40.5 \\
TSOD & $\times$     & $\checkmark$ & 45.4 / 66.6 & 43.4 / 64.6 & 32.2 / 50.4 & 16.5 / 29.5 \\
Ours & $\times$     & $\times$     & 45.0 / 67.1 & 42.6 / 64.4 & 33.7 / 53.4 & 21.6 / 36.7 \\
\bottomrule
\end{tabular}
\end{table}

\subsection{Ablation on Different Architecture}

\km{While our primary experiments focused on YOLOv5 and YOLOv11 following TSOD~\cite{zsqod}, the proposed method is not restricted to YOLO-based architectures and generalizes to other designs. To demonstrate this broader applicability, we conduct additional experiments using Mask R-CNN~\cite{he2017mask} with a Swin Transformer-T~\cite{liu2021swin} backbone, with results reported in Table~\ref{tab:transformer}. Unlike the single-stage YOLO models, Mask R-CNN follows a two-stage detection paradigm. In the first stage, a Region Proposal Network (RPN) predicts anchor objectness scores and bounding-box refinements to generate candidate region proposals. In the second stage, the RoI head performs per-proposal classification and bounding-box regression. Mask R-CNN further extends this with a parallel mask prediction branch. When applying Teacher-guided Adaptive Noise Reduction (TANR) during  QAT, the hard targets of classification-related branches are replaced using the corresponding teacher predictions (TANR) — specifically, the RPN objectness branch and the RoI classification branch. We note that this description is scoped to the classification-related components directly relevant to TANR.}

\subsection{Additional Examples of Synthesized Images}

\begin{figure}
    \centering
    \includegraphics[width=0.9\linewidth]{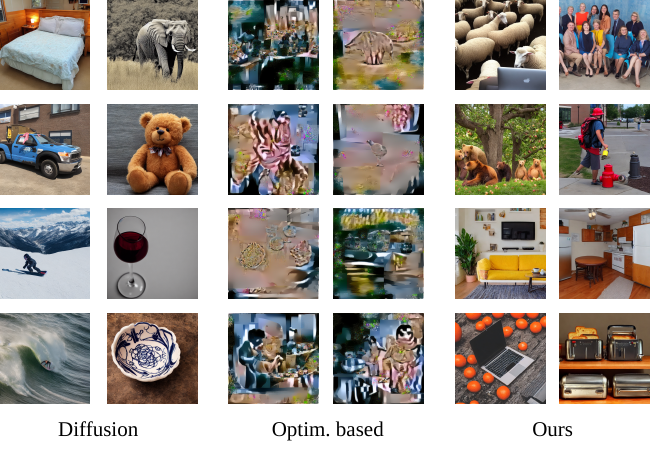}
    \caption{Additional examples of images generated using diffusion (GenQ), noise optimization (TSOD), and our method (GoodQ).}
    \label{fig:more_ex}
\end{figure}

\km{As an extension of the results shown in Figure~\ref{fig:teaser}, Figure~\ref{fig:more_ex} provides additional visual comparisons between the noise-optimized images (Optim. based), the images generated by naive diffusion (GenQ), and those generated by our proposed method (GoodQ).}

\subsection{Different Generator and Dataset}
In the main paper, GoodQ was primarily evaluated with the SD1.5 generator on the COCO dataset using a YOLO detector. In this section, we assess its generalization beyond this setting. Table~\ref{tab:goodq_generator} reports results when the generator is replaced with SD2.1, showing that GoodQ remains effective regardless of the underlying generator. Table~\ref{tab:goodq_dataset} further evaluates GoodQ on YOLOv11-s across the VOC~\cite{VOC} and HomeObjects-3K~\cite{HomeObjects} datasets. As the results indicate, GoodQ continues to perform consistently well, confirming that its effectiveness is not confined to a specific dataset.

\begin{table}[t]
\centering
\caption{Generalization of GoodQ to a different generator.
SD1.5 is replaced with SD2.1 on COCO. Each cell reports AP\,/\,AP$_{50}$.
Hyperparameters follow the SD1.5 setup. Best in \textbf{bold}.}
\label{tab:goodq_generator}
\setlength{\tabcolsep}{5pt}
\renewcommand{\arraystretch}{1.1}
\begin{tabular}{lcccc}
\toprule
\multirow{2}{*}{Method} &
\multicolumn{2}{c}{W3A3} & \multicolumn{2}{c}{W4A4} \\
\cmidrule(lr){2-3}\cmidrule(lr){4-5}
 & mAP & mAP50 & mAP & mAP${50}$ \\
\midrule
TSOD  & 5.0 & 9.9 & 18.4 & 32.3 \\
GoodQ & \textbf{8.7} & \textbf{17.1} & \textbf{20.5} & \textbf{35.3} \\
\bottomrule
\end{tabular}
\end{table}

\begin{table}[t]
\centering
\caption{Generalization of GoodQ across datasets.
Performance on YOLOv11-s for HomeObjects-3K and VOC.
Each cell reports mAP\,/\,mAP50. Hyperparameters follow the SD1.5 setup
except $\gamma{=}2$ (HomeObjects-3K) and $\gamma{=}0.5$ (VOC). Best in \textbf{bold}.}
\label{tab:goodq_dataset}
\setlength{\tabcolsep}{5pt}
\renewcommand{\arraystretch}{1.1}
\begin{tabular}{llcccc}
\toprule
\multirow{2}{*}{Dataset} & \multirow{2}{*}{Method} &
\multicolumn{2}{c}{W3A3} & \multicolumn{2}{c}{W4A4} \\
\cmidrule(lr){3-4}\cmidrule(lr){5-6}
 & & mAP & mAP50 & mAP & mAP50 \\
\midrule
\multirow{2}{*}{HomeObjects-3K} & TSOD  & 21.6 & 36.1 & 42.2 & 61.3 \\
                                & GoodQ & \textbf{25.4} & \textbf{40.9} & \textbf{42.6} & \textbf{64.1} \\
\midrule
\multirow{2}{*}{VOC}            & TSOD  & 8.5 & 18.8 & 42.0 & 63.9 \\
                                & GoodQ & \textbf{11.1} & \textbf{23.0} & \textbf{43.3} & \textbf{66.2} \\
\bottomrule
\end{tabular}
\end{table}

\end{document}